\documentclass[runningheads]{llncs}
\usepackage{graphicx}

\usepackage{tikz}
\usepackage{comment}
\usepackage{color}
\usepackage{subcaption}
\usepackage{wrapfig}
\usepackage{soul}
\usepackage{amsfonts}
\usepackage{amsmath,amssymb} 

\usepackage[numbers,sort,compress]{natbib}

\usepackage{colortbl}
\usepackage[accsupp]{axessibility}  

\usepackage[most]{tcolorbox}
\newtcolorbox[auto counter]{summary}[1][]{title={\bfseries Conclusion~\thetcbcounter},enhanced,drop shadow={black!50!white},
  coltitle=black,
  top=0.15in,
  attach boxed title to top left=
  {xshift=1.5em,yshift=-\tcboxedtitleheight/2},
  boxed title style={size=small,colback=pink},#1}

\usepackage{xcolor}
\usepackage{glossaries}
\usepackage{cleveref}
\usepackage{bbm}
\usepackage{paralist}
\usepackage{cancel}
\usepackage{multirow}
\usepackage{graphicx}
\usepackage{nicefrac}
\usepackage{booktabs}
\definecolor{red}{HTML}{E41A1C}
\definecolor{orange}{HTML}{FF7F00}
\definecolor{yellow}{HTML}{FFC020}
\definecolor{green}{HTML}{4DAF4A}
\definecolor{blue}{HTML}{377EB8}
\definecolor{purple}{HTML}{984EA3}

\Crefname{algocf}{Algorithm}{Algorithms}
\crefname{algorithm}{Algorithm}{Algorithms}
\crefname{figure}{Figure}{Figure}
\crefname{table}{Table}{Table}
\crefname{section}{§}{§§}
\Crefname{section}{§}{§§}
\crefformat{equation}{(#2#1#3)}
\crefname{property}{Property}{Property}
\crefname{theorem}{Theorem}{Theorem}
\crefname{proposition}{Proposition}{Proposition}
\crefname{lemma}{Lemma}{Lemma}

\glsdisablehyper{}
\newglossaryentry{lip}
{
    name=lipschitz,
    description={Lipschtiz constant}
}
\newglossaryentry{Lip}
{
    name=lipschitz,
    description={Lipschtiz constant}
}

\newacronym{nn}{NN}{Neural Networks}
\newacronym{sgd}{SGD}{stochastic gradient descent}
\newacronym{svd}{SVD}{Singular Value Decomposition}
\newacronym{sota}{SOTA}{state-of-the-art}
\newacronym{gan}{GAN}{Generative Adversarial Networks}
\newacronym{srn}{SRN}{Stable Rank Normalization}
\newacronym{sngan}{SN-GAN}{Spectral Normalization GAN}
\presetkeys{todonotes}{%
  backgroundcolor=blue!10!white,
  linecolor=blue!10!white,
  bordercolor=blue!10!white
}{}

\newcommand{\SKIP}[1]{}

\newcommand{\bfx}{\mathbf{x}}

\newcommand{\real}{\mathbb{R}}

\newcommand{\TPR}{\text{TPR}}
\newcommand{\FPR}{\text{FPR}}
\newcommand{\TNR}{\text{TNR}}
\newcommand{\FNR}{\text{FNR}}

\newcommand{\AUROC}{\text{AUROC}}

\begin{document}
\pagestyle{headings}
\mainmatter
\def\ECCVSubNumber{6459}  

\title{An Impartial Take to the CNN vs Transformer Robustness  Contest} 

\titlerunning{An Impartial Take to the CNN vs Transformer Robustness  Contest}
%
\author{Francesco Pinto\inst{1,2} \and
Philip H. S. Torr\inst{1} \and
Puneet K. Dokania\inst{1,2} \\
\email{\{francesco.pinto\}@eng.ox.ac.uk} 
}
\authorrunning{F. Pinto et al.}
%
\institute{$^1$University of Oxford \& $^2$Five AI Ltd., UK}
\maketitle

\begin{abstract}
Following the surge of popularity of Transformers in Computer Vision, several studies have attempted to determine whether they could be more robust to distribution shifts and provide better uncertainty estimates than Convolutional Neural Networks (CNNs). The almost unanimous conclusion is that they are, and it is often conjectured more or less explicitly that the reason  of this supposed superiority is to be attributed to the self-attention mechanism. In this paper we perform extensive empirical analyses showing that  recent state-of-the-art CNNs (particularly, ConvNeXt~\cite{ConvNeXt}) can be as robust and reliable or even sometimes more than the current state-of-the-art Transformers. However, there is no clear winner. Therefore, although it is tempting to state the definitive superiority of one family of architectures over another, they seem to enjoy similar extraordinary performances on a variety of tasks while also suffering from similar vulnerabilities such as texture, background, and simplicity biases.

\end{abstract}

\section{Introduction}
Transformers are a family of neural network architectures that became extremely popular in natural language processing, and are primarily  characterised by the extensive use of the attention mechanisms as defined in \citep{AttentionIsAllYouNeed}. 
Before Vision Transformers (ViT) \citep{ViT} were introduced, Transformers were considered difficult to use for computer vision applications due to the prohibitive computational complexity and memory requirements of the self-attention mechanism.
Since then, several transformer variants that are efficient to train with performance more competitive with the state-of-the-art CNNs like BiT \citep{BiT} (e.g. \citep{T2TViT, DeiT, SwinT}) have been proposed. 

The effectiveness of transformers compared to CNNs in computer vision applications has led to recent interest in comparing them in obtaining reliable predictive uncertainty and robustness to distribution shifts. The almost unanimous conclusion in the literature is that transformers exhibit: (1) better calibration \cite{TransformersCalibration}, (2) better robustness to covariate shift \citep{TransformersRobustLearners, DelvingDeepTransformers,AreTransformersMoreRobust, MorrisonDSTransformers}, and (3) better uncertainty estimation for tasks like out-of-distribution detection (OoD) \citep{LimitsOODFort,AreTransformersMoreRobust}. Currently, these conclusions are mostly misleading as (1) the recent convolutional architectures (ConNeXt) were not available for proper comparisons; (2) the comparisons are often performed with questionable assumptions (e.g. comparing model capacity solely based on their parameter count) or training procedures (e.g. trying to make the training as similar as possible for both the families at the cost of damaging the performance of either); and (3) the choice of the evaluation metrics is often not carefully justified and the most subtle aspects of the interpretation of the results were not identified.  Additionally, when it comes to explaining the outcome of the analysis, which mostly leads to concluding that Transformers are superior, the credit is often given (more or less explicitly) to the most prominent feature that distinguishes Transformers from CNNs: the self-attention mechanism. Yet, a fair comparison and an understanding of whether and how self-attention modules would allow learning superior features compared to convolutional models is needed before providing a definitive answer regarding the superiority of one over another.

Taking a step in this direction, we thoroughly evaluate the robustness and reliability of most recent state-of-the-art Transformers (ViT \citep{ViT} and SwinT \citep{SwinT}) and CNN architectures (BiT \citep{BiT} and ConvNeXt \citep{ConvNeXt}) on ImageNet-1K \citep{Imagenet1K}. We would like to highlight that we do not modify the training recipes of CNNs and Transformers to ensure that they are at their current best during comparisons. The main takeaways of our work are:
\begin{enumerate}
    \item \textbf{Simplicity bias experiment}~\cite{SimplicityBias}. Transformers, just like CNNs, also suffer from the so-called simplicity bias. They are somewhat similar to CNNs in finding shortcuts (undesirable) to solve the desired task. Therefore, as opposed to the common notion, despite the capability of the self-attention modules to communicate globally, Transformers as well tend to focus on easy-to-discriminate parts of the input and conveniently ignore other complex-yet-discriminative ones. Hence, similar to CNNs, they might just be learning to combine sets of simple and potentially spurious features, rather than more complex and invariant ones. Based on this experiment, we discourage the common trend in the literature to give unnecessary praise to the self-attention module of Transformers anytime these perform better against CNNs. More theoretical developments, analyses, and well-thought experiments are needed to support such claims.
    \item We show that for out-of-distribution detection task, CNNs and Transformers \textbf{perform equally well}.  
    We also highlight why, unless domain-specific assumptions are made, preferring AURP over AUROC in situations of data imbalance (which generally is the case) might give the false impression of one model being significantly superior to others. 
    \item In-distribution calibration of the best performing CNN model (in terms of accuracy) is better than the best performing Transformer. However, there is \textbf{no clear winner} that performs the best in all the experiments including covariate shift. 
    \item Again, there is \textbf{no clear winner} in detecting misclassified inputs.
\end{enumerate}

These takeaways also suggest that the inductive biases induced in CNNs by using the design components popularised by Transformers (e.g. GeLU \citep{GeLU} activations, LN normalization \citep{LNorm} etc.), but without using the self-attention mechanism, might be highly effective in bridging the gap between the two in terms of robustness. However, this speculation requires further analysis as there are too many variables involved in designing a model (from architectural design choices to optimization algorithms) and the interplay between them is not well understood yet. 

\section{Experimental Design and Choices}

\subsection{Setup} 
\noindent {\bf Models.} We consider state-of-the-art convolutional and non-convolutional models for our analysis. 
\begin{enumerate}
    \item {\bf BiT}~\citep{BiT}: It is a very commonly used family of fully convolutional architectures. Its members are ResNet variants that have been shown to achieve state-of-the-art accuracy on ImageNet classification and that, with an appropriate fine-tuning procedure, transfer well to many other datasets. In this paper we consider BiT-R50x1, BiT-R50x3, BiT-R101x1, BiT-R101x3, BiT-R152x2, BiT-R152x4 (where R50/101/152 indicates the ResNet variant, and the multiplicative factor scales the number of channels).
    \item {\bf ConvNeXt}~\citep{ConvNeXt}: A recent family of fully convolutional architecture that is very close to the non-convolutional Transformer models in terms of training recipes and design choices. Its members have been shown to produce either comparable or superior performance to Transformers on several large-scale datasets. ConvNeXt exemplifies how advancing state-of-the-art in one family of networks can yield architecture design choices that, if adapted properly, can benefit other families of networks too. Our conclusions heavily rely on the careful architecture design process of ConvNeXt. We consider ConvNeXt-B, ConvNeXt-L, ConvNeXt-XL variants. Here and also for other models, B, L and XL indicate the capacity (B = Base, L = Large, XL = Extra Large).
    \item {\bf ViT} \citep{ViT}: First successful use of Transformers on vision tasks. Its members still exhibit state-of-the-art performances. We consider ViT-B/16 and ViT-L/16\footnote{We omit ViT-B/32 ViT-L/32 as we find them to always underperform with respect to ViT-B/16 and ViT-L/16 (a similar observation was made in \citep{TransformersRobustLearners}). Similarly, we also omit DeiT \citep{DeiT} as it underperforms compared to SwinTransformers.}, where 16 indicates the input token patch size.
    \item {\bf SwinTransformer}  \citep{SwinT}: A family of transformers implementing a hierarchical architecture employing a shifting window mechanism. We consider the  Swin-B and Swin-L variants. We use patch size of 4 pixels and shifted windows of size 7 as they provide highly competitive performance.
\end{enumerate}

\noindent {\bf Training.} Unless stated otherwise, all the considered architectures have been pre-trained on ImageNet-21k \citep{Imagenet21K} and fine-tuned on ImageNet-1k \citep{Imagenet1K}. We use the trained checkpoints available in the \texttt{timm library} \citep{timm} except only for the simplicity bias experiments where we fine-tune the models on our own. Additional results showing the impact of pre-training are shown in Appendix~\ref{sec:additionalResults}.

\noindent {\bf Datasets.} Since the in-distribution dataset is ImageNet-1K, we use ImageNet-A \citep{NaturalAdversarialExamples}, ImageNet-R \citep{ImageNetR}, ImageNetv2 \citep{ImageNetv2}, ImageNet-Sketch \citep{ImageNet-Sk} for the \textit{domain-shift} experiments. For \textit{out-of-distribution} detection experiments, we use ImageNet-O \citep{NaturalAdversarialExamples}. For our preliminary analyses to understand \textit{existing biases} in Transformers and CNNs, we use ImageNet9 \citep{ImageNet-9}, the Cue-Conflict Stimuli dataset \citep{ImageNet-Sk}, and also \textit{synthesize} a dataset by combining MNIST and CIFAR-10 datasets.
For Imagenet experiments, we apply the standard preprocessing pipeline. Additional results showing the impact of input preprocessing are shown in Appendix~\ref{sec:additionalResults}.

\subsection{Yet Another Analysis?}
Before we begin discussing our analyses, we would like to mention how we differ from the existing ones.

Closest to ours is a recent analysis presented by~\cite{AreTransformersMoreRobust} which involves rather simpler architectures for both Transformers (DeiT) and CNNs (ResNet-50), and also drops transformer-specific training techniques (for instance, reducing training epochs to 100 from 300, removing augmentations and regularisation techniques etc.). This indeed brings DeiT down to CNNs in terms of training procedure, however, makes DeiT underperform significantly. Although they derive interesting insights, the applicability of these insights for a practitioner with an intent to identify the most robust and best performing model is somehow limited. Therefore, we not only consider a wider variety of CNNs and Tranformers in our analysis, we also do not modify their standard training recipes so that their best performance is being compared. In ~\cite{OursAreTransformersMore}, authors do  provide a partial and preliminary analysis questioning the existing literature, however, solid evidence is still lacking. Another work~\cite{robustART} showed superiority of CNNs over Transformers on natural covariate-shift datasets. Differently from them, our analysis not only considers these metrics, but also the performance in terms of calibration, misclassification detection, and out-of-distribution detection. Other recent work~\cite{MorrisonDSTransformers, DelvingDeepTransformers} performs partially overlapping analyses reaching the same conclusion about the superiority of Transformers. However, \citep{MorrisonDSTransformers} do not consider recent CNN models, and also compare Transformers pre-trained on ImageNet-21K with CNNs that are trained from scratch on ImageNet-1K. Instead, \citep{DelvingDeepTransformers} only compares with the extremely simple CNN variants.

We would also like to highlight that comparing different models based on their capacity (determined solely based on their number of parameters) might lead to wrong conclusions. How well a model would preform in practice is heavily dependent on the nature and the composition (hierarchy, depth etc.) of the underlying functions, not just on the number of parameters. To provide a widely known example, an MLP with one hidden layer and enough hidden units (large number of parameters) can theoretically fit most functions of interest, and it is known to be a universal function approximator~\cite{HORNIK1989359, Cybenko_superpositionSigmoidal}. However, in practice, they underperform compared to a deep network (with same or even less number of parameters). The interaction of inductive biases and training procedures plays an important role towards finding solutions that generalise well. 

Therefore, although the number of parameters can be a proxy for comparing model capacity, in practice, it can be misleading. Indeed, when compute and memory constraints are imposed, a practitioner will always find the best performing model satisfying such constraints rather than choosing a model based on the parameter count\footnote{Consider that ViT-L/32 has about 307M parameters, ViT-L/16 has 305M, yet ViT-L/32 requires about 15GFLOPS, while ViT-L/16 requires about 61GFLOPS, and ViT-L/32 exhibits lower accuracy and robustness than ViT-B/32 \citep{TransformersRobustLearners}}. We provide discussions and empirical findings (using standard complexity measures) to support our arguments above in Appendix~\ref{sec:capacity}. 
\section{Empirical Evaluation and Analysis}

\subsection{Are Transformer Features More Robust than CNN ones?}
\label{sec:bias_experiments}
There is no clear answer to this question in the literature. It is known that for a model to generalise to previously unseen domains, its predictions should not depend on spurious features that are specific to the distribution from which the training and test in-domain sets are sampled from, but on robust features that generalise across other domains under covariate-shift \citep{DataShiftBook}. Typical examples of spurious features described in literature are the background's colour, textures and generally any simple pattern that correlates strongly with the labels in the training set but not in the test set \citep{IRM}. 

It is usually conjectured in the literature that Transformers might be learning more robust features than CNNs because of the ability of their self-attention modules to communicate globally within a given input \citep{TransformersRobustLearners}. Which, in fact, is equivalent to implicitly criticizing the convolutional inductive biases of CNNs for their relatively poor robustness. 
Before we begin comparing these two families in terms of robustness, here we first present a few experiments to analyse their vulnerabilities. These experiments show that \emph{the sole presence of the self-attention mechanism is not sufficient for Transformers to neglect spurious features}, and they result to be as biased as CNNs towards them. 

\begin{table}
		\centering 
\resizebox{\textwidth}{!}{
	\begin{tabular}{cc|ccc|ccc|c|c}
	
	   \toprule 
		 & & \multicolumn{3}{c|}{\textbf{SB}}  & \multicolumn{4}{c|}{\textbf{BB}} &
		 \multicolumn{1}{c}{\textbf{TB}} \\

& $\#$ params (M) & \textbf{In-domain}  & \textbf{R-MNIST} & \textbf{R-CIFAR}  & \textbf{O} ($\uparrow$) & \textbf{MS} ($\uparrow$) & \textbf{MR} ($\uparrow$) & \textbf{BG-Gap} ($\downarrow$) & \textbf{CCS}($\uparrow$) \\
\hline

BiT-R50$\times$1    &25& 100 & 48.39 & 100 &       94.57  &  83.21 & 76.2 &   7.00 & 31.09    \\
BiT-R50$\times$3    &217&100& 48.14 &100&       95.14 &   85.14 & 80.22 &  4.92 & 33.12    \\
BiT-R101$\times$1   &44&100&48.50&99.94&       94.17  &  81.28 & 75.19 &  6.09 & 32.81    \\
BiT-R101$\times$3   &387&100&48.19&99.89&       94.32  &  81.19 & 76.67 &  4.52 & 32.58    \\
BiT-R152$\times$2  &232&100 & 48.39 & 99.94 &       94.64  &  80.05 & 75.09 &  4.95 &  35.47   \\
BiT-R152$\times$4   &936&100& 48.19 & 100 & 95.01 & 81.16 & 75.33 & 5.83 & 37.19\\ 
\hline

ConvNeXt-B    &88& 100 & 48.29  & 99.94  &       97.95 &   93.95 & 90.42 &  3.53 & 30.63     \\
ConvNeXt-L    &196&100& 48.20& 99.89&         98.2   &  95.19 & 91.63 &  3.56 & 35.16    \\
ConvNeXt-XL   &348&100& 48.75 & 99.69&         98.49  &  95.23 & 92.3 &   2.93 & 36.95     \\
\hline
\hline

ViT-B/16     &86& 100 & 48.59 & 99.79   &       97.36  &  92.35 & 88 &     4.34 & 30.78    \\
ViT-L/16     &304& 100 & 52.79 & 95.66   &       98.02  &  94.05 & 90.05 &  4  & 47.19   \\
\hline

Swin-B     &87& 100 & 48.75 & 99.64    &       97.75  &  90.94 & 86.47 &  4.47   & 26.95  \\
Swin-L     &195& 100 & 48.69 & 99.74   &       98.02  &  92.99 & 88.47 &  4.52    &  30.08\\

\bottomrule
	\end{tabular}%

    }
    \caption{\textbf{Simplicity bias (SB), Background bias (BB) and Texture bias (TB) experiments}. For \textbf{SB}, in-domain indicates the accuracy when MNIST and CIFAR images are associated as in the training set. \textit{A model suffers from SB if R-MNIST accuracy is close to random whereas R-CIFAR accuracy is close to the in-domain}. For \textbf{BB}, we report the absolute accuracy on the original (O), mixed-same (MS), and mixed-random (MR) datasets, respectively.  \textbf{BG-Gap}  defined as the difference in accuracy between MS and MR, quantifies the impact of background in producing correct classifications. For \textbf{TB}  we report the CCS accuracy. All quantities in the table are percentages (\%). 
}
    \label{tab:i9_bkg}
\end{table}

\begin{figure}[ht]
  \subfloat[\textbf{Cars}, target label \textbf{-1}
  ]{
	\begin{minipage}[c][1\width]{
	   0.3\textwidth}
	   \centering
	   \includegraphics[width=1\textwidth]{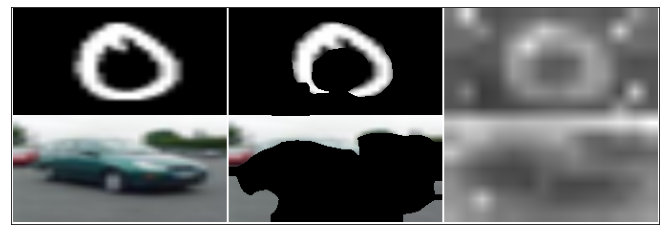}
	   \includegraphics[width=1\textwidth]{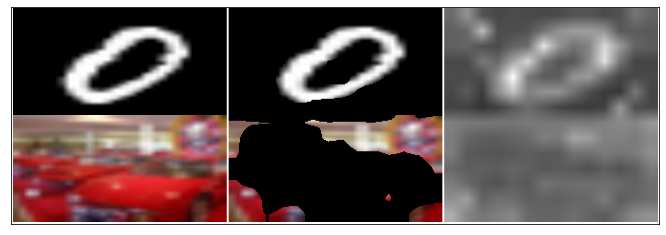}
	   \includegraphics[width=1\textwidth]{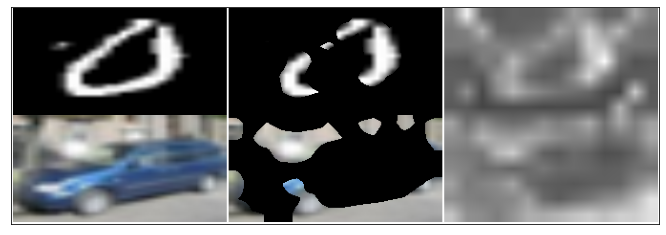}

	\end{minipage}}
 \hfill 	
  \subfloat[\textbf{Trucks}, target label \textbf{1}]{
	\begin{minipage}[c][1\width]{
	   0.3\textwidth}
	   \centering
	   \includegraphics[width=1\textwidth]{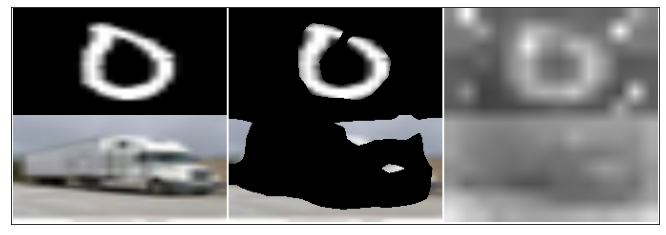}
	   	   \includegraphics[width=1\textwidth]{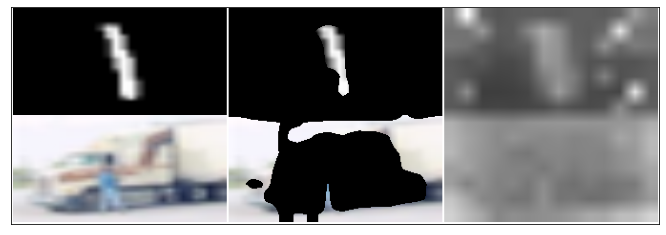}
	   \includegraphics[width=1\textwidth]{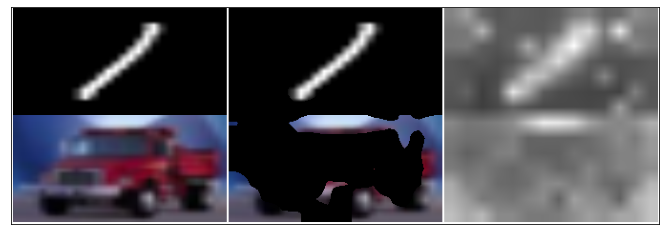}

	\end{minipage}}
 \hfill	
  \subfloat[Top to bottom: attention at layer 1, 4 and 12.]{
	\begin{minipage}[c][1\width]{
	   0.3\textwidth}
	   \centering
  	   \includegraphics[width=1\textwidth]{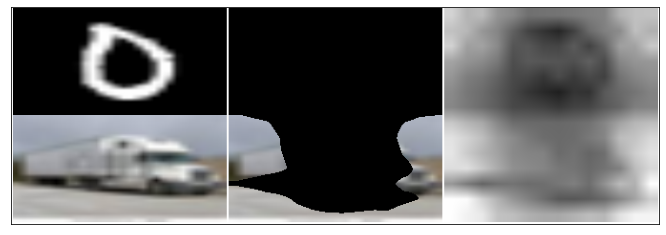}

	   \includegraphics[width=1\textwidth]{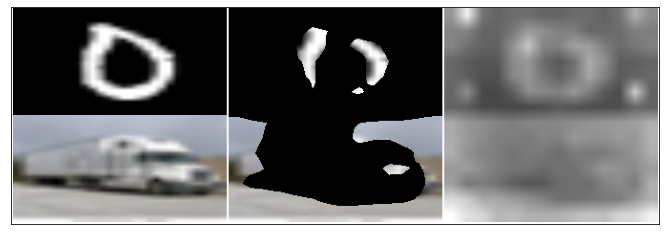}
	   \includegraphics[width=1\textwidth]{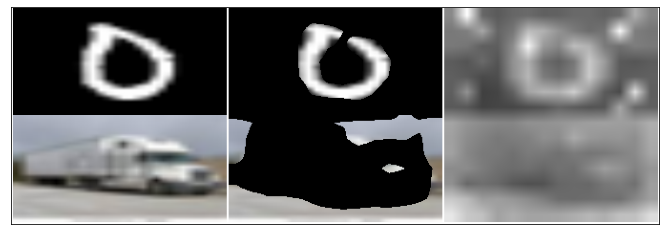}

	\end{minipage}}

    \caption{\textbf{Simplicity Bias Experiment for Transformers}: For each triplet of images, from left to right:  input image, test image without pixels on which the attention value is below the 70\% quantile and the attention map visualization. The attention maps show that the Transformer (ViT-B/16) gives high attention values to simple features and neglect the complex ones. 
      }
    \label{fig:simplicity_bias_attention}
\end{figure}

{\bf Simplicity Bias Experiment.} The intent of this experiment is to understand what Transformers and CNNs prefer to learn in situations where it is possible to focus only on the simple discriminative features of the input and ignore the complex discriminative ones in order to perform well on the task. This experiment was proposed and analysed on CNNs by~\citep{SimplicityBias}. Following their work, we first create a binary classification task where the input $X = [\bfx, \bar{\bfx}]$ is composed of the concatenation of $\bfx$ and $\bar{\bfx}$,  both discriminative, and learning features for \textit{either or both} will lead to an accurate classifier. We design this task such that, say, $\bar{\bfx}$ is more complex\footnote{We understand that defining complexity is subjective. Here we assume that something that is visually more complex (having more colors, shapes, textures etc.) across the training set would require learning more complex features.} than $\bfx$.  Then, under this setting, a trained classifier suffers from simplicity bias if (1) fixing $\bfx$ and randomly modifying $\bar{\bfx}$ in the input does not change its prediction, and (2) fixing $\bar{\bfx}$ and randomly modifying $\bfx$ in the input drops the test accuracy to the random prediction baseline. 

To create the dataset for the above experiment, $\bfx$ is taken from the MNIST dataset \citep{MNIST} (randomly sampled image of a certain digit) while $\bar{\bfx}$ from the relatively more complex CIFAR-10 (randomly sampled image of a certain label). For instance, say digit \textbf{0} is associated to \textbf{car} and the whole concatenated image is assigned label $+1$, and digit \textbf{1} is associated to \textbf{truck} and the concatenated image is labelled $-1$. Refer to the top left of Figure~\ref{fig:simplicity_bias_attention}. During training, this relationship holds true for all the examples (in-domain). We fine-tune our classifiers on this dataset for 3 epochs (it is easy to converge on this dataset). At test time, we either randomise the MNIST part of the image (R-MNIST) or the CIFAR part of the image (R-CIFAR) for the analysis. Results are  reported in Table~\ref{tab:i9_bkg}.

As it can be seen, the accuracy is almost the same for all the models (except in ViT-L/16) even if the CIFAR (more complex) part of the input is completely randomized (R-CIFAR). However, the accuracy drops to nearly random ($50\%$) when the MNIST part of the input is randomized (R-MNIST). This shows that both families, Transformers and CNNs, rely on MNIST for classification and are agnostic to the CIFAR component. Hence, both are prone to simplicity bias. To understand which are the most prominent features leveraged by the Transformer, we visualize the pixels that fall above the $70\%$ quantile of the intensity values in the attention map, and blacken the ones that fall below it in Figure \ref{fig:simplicity_bias_attention}. This figure confirms that Transformer's self-attention mechanism neglects complex features in favour of simple features. Figure \ref{fig:simplicity_bias_attention} (c) also shows how the self-attention changes through the layers of the transformer. At the first layer there is no specific focus on the MNIST digit, but as the layers progress (i.e. as the features specialise to be useful for the classification), the attention values increase around the digit.

{\bf Reliance on Backgrounds and Texture.} Here we measure the performance of  several architectures on a benchmark that measures the reliance of features on backgrounds and textures: ImageNet9~\citep{ImageNet-9} and the Cue-Conflict Stimuli~\citep{ImageNet-Sk}. 

The {\bf ImageNet9} dataset selects a subset of labels and images from the original ImageNet dataset. In our experiments we measure the accuracy on the full images of the dataset (\emph{original split}), images in which the background has been swapped with another image of same class (\emph{mixed-same}), images in which the background has been swapped with another image of different class chosen at random (\emph{mixed-random}). Sample images are provided in Appendix~\ref{sec:sampleimages}. The authors of this dataset suggest taking the gap between the accuracy on mixed-same and mixed-random as a quantifier of the reliance on background information to produce accurate predictions. As it can be seen from Table \ref{tab:i9_bkg}, some of the highest capacity BiT models do not rely more on the background than ViT-B/16, SWIN-B and SWIN-L. ConvNeXt models rely on backgrounds even less than Transformers, suggesting that the self-attention mechanism might not be the only factor responsible for the difference observed between low-capacity ResNets and Transformers. 

The {\bf Cue-Conflict Stimuli} dataset alters the texture information of an image using style transfer: given an image of a certain class, it uses as style-image a sample from another class (sample images in Appendix~\ref{sec:sampleimages}). The purpose is to deceive classifiers that overly rely on textures to make predictions.  As it can be seen in Table  \ref{tab:i9_bkg}, although the top performing model is ViT-L/16 (with a significant margin), Swin Transformers exhibit an even heavier reliance on texture than ConvNeXt models, and ViT-B/16 performs comparably to ConvNeXt-B. This suggests that the sole presence of the self-attention in an architecture is not sufficient for the model to not be biased towards texture information. 
\begin{summary}[title style={colback=grey},colback=white]
\begin{itemize}
    \item[$\circ$] Transformers can leverage spurious features just like CNNs. They can be comparably prone to various biases such as simplicity bias, background bias, and texture bias. The sole presence of self-attention might not be sufficient to avoid such biases.
\end{itemize}
\end{summary}

\subsection{Out-of-Distribution Detection}
\label{sec:ood_sec}
Current notion in the literature is that Transformers are better than CNNs at detecting OoD samples~\cite{TransformersRobustLearners}. 

We compare various CNN and Transformer models at the task of detecting ImageNet-O samples from ImageNet-1K. ImageNet-O contains 2K samples in 200 classes, while the subset of ImageNet-1K used as the corresponding in-distribution set contains 10K samples \citep{NaturalAdversarialExamples} (therefore, there is a stark imabalance in the number of samples belonging to the two sets). Both ImageNet-O and ImageNet-1K (test) samples are fed to the classifier, for each point an uncertainty score is computed and a binary threshold-based classifier is used to distinguish between them. Since the choice of the threshold depends on the risk exposure desired for a certain application, a standard evaluation procedure considers all the risk thresholds and computes the AUROC (Area Under the Receiver Operating Characteristic curve) and the AUPR (Area Under the Precision-Recall curve).
\begin{table*}
\centering
\resizebox{\linewidth}{!}
{%
	\begin{tabular}{c|cc|cc|cc|cc}
		\toprule
		
		& \multicolumn{4}{c|}{IND=1,OoD=0} & \multicolumn{4}{c}{IND=0,OoD=1} \\
		& \multicolumn{2}{c}{\textbf{Imbalanced}} & \multicolumn{2}{c|}{\textbf{Balanced}}  & \multicolumn{2}{c}{\textbf{Imbalanced}} & \multicolumn{2}{c}{\textbf{Balanced}} \\

			 &	 AUROC ($\uparrow$) & AUPR ($\uparrow$) & AUROC ($\uparrow$) & AUPR ($\uparrow$)  &	 AUROC ($\uparrow$) & AUPR ($\uparrow$) & AUROC ($\uparrow$) & AUPR ($\uparrow$)\\
\hline 
BiT-R50x1 & 65.17 & 90.15 &65.17 & 65.81 &65.17 & 23.30 &65.17 & 60.13 \\
BiT-R50x3 & 74.56 & 92.30 &74.56 & 71.28 &74.56 & 36.26 &74.56 & 72.49 \\
BiT-R101x1 & 70.34 & 91.35 &70.34 & 68.75 &70.34 & 28.53 &70.34 & 66.11 \\
BiT-R101x3 & 77.32 & 93.40 &77.32 & 74.84 &77.32 & 38.74 &77.32 & 74.66 \\
BiT-R152x2 & 77.46 & 93.51 &77.46 & 75.23 &77.46 & 38.24 &77.46 & 74.43 \\
BiT-R152x4 & 80.07 & 94.39 &80.07 & 78.10 &80.07 & 44.25 &80.07 & 78.17 \\
\hline 
ConvNeXt-B & 85.72 & 95.53 &85.72 & 81.74 &85.72 & 59.15 &85.72 & 85.53 \\
ConvNeXt-L & 89.07 & 96.90 &89.07 & 86.96 &89.07 & 65.33 &89.07 & 88.55 \\
ConvNeXt-XL & \underline{90.04} & 97.19 & \underline{90.04} & 88.11 &\underline{90.04} & 68.50 &\underline{90.04} & 89.75 \\
\hline 
\hline 
 
ViT-B/16 & 79.89 & 95.26 &79.89 & 82.30 &79.89 & 36.77 &79.89 & 73.77 \\
ViT-L/16 & \textbf{90.60} & 97.85 & \textbf{90.60} & 91.27 & \textbf{90.60} & 64.58 & \textbf{90.60} & 88.90 \\
\hline 
Swin-B & 83.74 & 95.29 &83.74 & 81.01 &83.74 & 52.93 &83.74 & 82.80 \\

Swin-L & 87.76 & 96.55 &87.76 & 85.67 &87.76 & 62.51 &87.76 & 87.27 \\

        \bottomrule
	\end{tabular}%
}
\caption{ImageNet-O: \textbf{OoD} performance analysis when in-distribution samples are assigned label 1 and  OoD label 0, and vice-versa (with and without rebalancing). AUROC (\%) is invariant whereas AUPR (\%) is extremely sensitive to these design choices. The best performing method based on AUROC is in bold and the second best is underlined. The gap between the two is marginal.}
	\label{tab:ood}
\end{table*}

\noindent {\bf AUPR vs AUROC?} We start by observing that the apparent complexity in distinguishing ImageNet-O samples from ImageNet-1K observed in the literature (e.g. \citep{NaturalAdversarialExamples, TransformersRobustLearners}) mostly depends on the interaction between specific evaluation choices. The AUPR, in the case of an imbalanced number of samples belonging to the positive and negative classes, is known to prefer one class over another. However, for out-of-distribution evaluation, unless additional domain-specific assumptions are made, there is no preferred mistake: confusing an in-distribution sample with an out-of-distribution sample or viceversa are both equally important mistakes.  To exemplify why the AUPR can yield misleading conclusions, in Table \ref{tab:ood} we consider different possible assignments of the positive class and apply a rebalancing technique as well. Recent work~\citep{NaturalAdversarialExamples, TransformersRobustLearners} concluding that there exist a dramatic gap between CNNs and Transformers on OoD detection performance report values when OoD samples are considered as positives (third column from the right). In this setting, for instance, the performance of BiT-R50x1 is less than half of the performance of ViT-L/16, and extremely low (with respect to the attainable maximum of 100). However, only rebalancing the number of samples\footnote{We oversample OoD samples ($4\times$) so that both in-distribution and OoD datasets have 10000 samples each. We could rebalance them also by randomly sampling 2000 out of the 10000 in-distribution samples, but this could induce some variance in the metrics; we also observed that the average of this strategy coincides with the balancing strategy.}  the performance of BiT-R50x1 rises to more than two thirds of the performance of ViT-L/16 (last column on the right). Alternatively, if the choice of the positive and negative class is flipped, in an imbalance condition, one can obtain an absolute gap between the performance of BiT-R50x1 and ViT-L/16 of less than 8\% (third column from the left). If one drew conclusions solely based on this column, one would think there is only a marginal difference between the performance of the two models. This gap widens when rebalancing the number of samples (fourth column from the left). This exemplifies how widely the AUPR can vary based on evaluation choices that, in the lack of domain-specific assumptions, are arbitrary.
On the other hand, the AUROC does not vary across all the considered evaluation setups, because it gives the same importance to both types of errors that can occur.   
These results allow us to conclude that, for the considered models, ImageNet-O is evidently not as hard to distinguish from ImageNet as it is believed to be. 

For completeness, in Appendix~\ref{sec:auroc_invariant} we provide a proof to show that AUROC is invariant to the choice of positive and negative classes.

{\bf Comparing Transformers and CNNs} From the AUROC values in Table \ref{tab:ood} it is clear that the top-performing CNN (ConvNeXt-XL) is competitive to the top-performing Transformer (ViT-L/16). ConvNeXt-L outperforms Swin-L, and ConvNeXt-B outperforms Swin-B. The best performing BiT (BiT-R152$\times$4) outperforms ViT-B.

\begin{summary}[title style={colback=grey},colback=white]
\begin{itemize}
    \item[$\circ$] CNNs can perform as well as Transformers for OoD detection.
    \item[$\circ$] With no domain-specific assumptions regarding the importance of one category over another (in-distribution vs OoD), AUROC should be preferred over AUPR as it is stable across evaluation choices.
\end{itemize}

\end{summary}

\subsection{Calibration on In-Distribution and Domain-Shift}
\label{sec:calibration}
A model is said to be calibrated if its confidence (i.e. the maximum probability score of the softmax output) and its accuracy match. The idea is to attribute to the confidence the frequentist probabilistic meaning of counting the amount of times the model is correct.  
Several measures have been proposed targeted specifically towards quantifying the said mismatch between a classifier's confidence and its accuracy. These measures are primarily the variants of the well-known Expected Calibration Error (ECE) \citep{Naeini2015-ECE} such as the recently proposed Adaptive Calibration Error (AdaECE) \citep{mukhoti2020calibrating}.

{\bf Comparing Transformers and CNNs}
On in-domain data (Table \ref{tab:imagenet1k_clean}), ViTs produce the lowest calibration error and Swin transformers are outperformed by ConvNeXts.  On covariate-shifted inputs (Table \ref{tab:imagenet1k_ds}), ViTs produce higher calibration error than ConvNeXts and Swin transformers, and the model producing the lowest calibration error is the Swin-L. Consistently with \citep{TransformersCalibration}, within a family of models, the ECE typically decreases as the number of parameters (and also the accuracy) increases. 

	\begin{table}
		\centering
\resizebox{0.5\textwidth}{!}{

	\begin{tabular}{c|ccc}
		\toprule

		& \multicolumn{3}{c}{\textbf{ImageNet-1K (Test)}}\\

		 & \textbf{Acc} ($\uparrow$) & \textbf{ECE} ($\downarrow$) & \textbf{AdaECE} ($\downarrow$)\\

\hline
BiT-R50x1       & 74.03  & 3.49 & 3.45 \\ 
BiT-R50x3       & 77.92  & 6.56 & 6.51 \\ 
BiT-R101x1      & 75.85  & 5.10 & 5.10 \\ 
BiT-R101x3      & 78.20  & 7.63 & 7.63 \\ 
BiT-R152x2      & 78.00  & 6.37 & 6.37 \\ 
BiT-R152x4      & 78.16  & 9.38 & 9.38 \\ 
\hline                                
                                      
ConvNeXt-B  & 85.53 & 2.87 & 2.82  \\ 
ConvNeXt-L  & 86.29 & 2.27 & 2.34  \\ 
ConvNeXt-XL & \textbf{86.58} & 2.38 & 2.29  \\ 
\hline                                
\hline                                
                                      
ViT-B/16     & 78.01  & \textbf{1.40} & \textbf{1.41}  \\
ViT-L/16     & 84.38  & 1.81 & 1.83  \\
                                      
\hline

SWIN-B & 84.71 & 8.40 & 8.40   \\     
SWIN-L & 85.83 & 5.50 & 5.50   \\     





\bottomrule
	\end{tabular}%

    }
    \caption{\textbf{In-distribution} accuracy (\%) and \textbf{calibration} (\%) for ImageNet-1K.
}
    \label{tab:imagenet1k_clean}
\end{table}

	\begin{table}
		\centering
\resizebox{\textwidth}{!}{

	\begin{tabular}{c|ccc|ccc|ccc|ccc}
		\toprule
		 & \multicolumn{12}{c}{\textbf{Domain-Shift}}  \\

		&   \multicolumn{3}{c|}{\textbf{ImageNet-R}} &
		\multicolumn{3}{c|}{\textbf{ImageNet-A}} &
		\multicolumn{3}{c|}{\textbf{ImageNet-V2}} &
		 \multicolumn{3}{c}{\textbf{ImageNet-SK}}\\

		 & \textbf{Acc} ($\uparrow$) & \textbf{ECE} ($\downarrow$) & \textbf{AdaECE} ($\downarrow$)  & \textbf{Acc} ($\uparrow$) & \textbf{ECE} ($\downarrow$) & \textbf{AdaECE} ($\downarrow$) & \textbf{Acc} ($\uparrow$)&  \textbf{ECE} ($\downarrow$) & \textbf{AdaECE} ($\downarrow$) & \textbf{Acc} ($\uparrow$)&  \textbf{ECE} ($\downarrow$) & \textbf{AdaECE} ($\downarrow$)    \\

\hline
BiT-R50x1      & 39.87 & 15.50 & 15.50 & 10.97 & 42.94 & 42.94 & 62.70 & 8.49 & 8.45   & 27.34  & 24.87 & 24.87  \\
BiT-R50x3      & 46.39 & 14.65 & 14.65 & 24.08 & 34.48 & 34.48 & 66.36 & 13.13 & 13.13 & 33.47  & 28.55 & 28.55  \\
BiT-R101x1     & 41.72 & 12.24 & 12.24 & 16.29 & 36.68 & 36.68 & 64.61 & 10.21 & 10.21 & 28.69  & 24.37 & 24.37  \\
BiT-R101x3     & 47.00 & 15.80 & 15.80 & 27.11 & 32.92 & 32.92 & 66.44 & 14.44 & 14.39 & 34.15  & 30.67 & 30.67  \\
BiT-R152x2     & 48.02 & 15.38 & 15.38 & 27.15 & 32.25 & 32.25 & 66.76 & 12.14 & 12.13 & 35.70  & 28.41 & 28.41  \\
BiT-R152x4     & 47.57 & 15.32 & 15.32 & 30.84 & 29.93 & 29.93 & 67.12 & 15.75 & 15.67 & 35.08  & 31.45 & 31.45  \\
\hline

ConvNeXt-B    & 62.46 & 2.57 & 2.51  & 52.63 & 8.28 & 8.31  & 75.43  & \textbf{2.91} & \textbf{2.78} & 48.64    & 8.85 & 8.84     \\
ConvNeXt-L    & 64.57 & 3.00 & 3.08  & 58.23 & 7.57 & 7.26  & 76.77 & 3.72 & 3.85    & 50.08   & 10.31 & 10.31   \\
ConvNeXt-XL   & \textbf{66.01} & 2.92 & 2.90  & \textbf{61.11} & 7.54 & 7.21  & \textbf{77.20} & 4.00 & 4.24    & \textbf{52.67}   & 11.16 & 11.15   \\
\hline
\hline

ViT-B16     & 43.15 & 5.21 & 5.21 & 24.17 & 22.89 & 22.89 &  66.25 & 4.71 & 4.68 & 18.18 & 13.02 & 13.02 \\
ViT-L16     & 61.54 & 3.07 & 3.07 & 47.08 & 11.99 & 11.99 &  74.28 & 5.34 & 5.22 & 45.96 & 10.67 & 10.67   \\

\hline

Swin-B      & 59.63 & 2.18 & 2.17 & 49.72 & 8.77 & 8.76 & 74.74 & 4.92 & 4.81 & 45.07 & \textbf{7.75} & \textbf{7.75} \\
Swin-L      & 64.24 & \textbf{2.14} & \textbf{2.11} & 59.52 & \textbf{6.19} & \textbf{6.33} & 76.65 & 3.03 & 3.14 & 48.87 & 8.72 & 8.71 \\





\bottomrule
	\end{tabular}%

    }
    \caption{\textbf{Domain-shift} accuracy (\%) and \textbf{calibration} (\%) for ImageNet-1K.
}
    \label{tab:imagenet1k_ds}
\end{table}

\begin{summary}[title style={colback=grey},colback=white]
\begin{itemize}
    \item[$\circ$] There is no one model that performs the best in all the covariate shift experiments in terms of calibration. Transformers or CNNs either can be better or worse depending on the experiment. 
    \item[$\circ$] The best performing model in terms of accuracy is not the most calibrated one.
\end{itemize}
\end{summary}

{\bf Is Low Calibration Error Enough for a Classifier to be Reliable?} A perfectly calibrated classifier can still be highly inaccurate and unreliable. For example, consider the binary case where there are $70$ negative test samples and $30$ positives. A classifier that has learned to classify every sample to a negative class with a confidence of $0.7$ will be perfectly calibrated, however, only $70\%$ accurate. Since neural networks trained on cross-entropy loss are known to be overconfident~\cite{GuoCalibration}, even if we somehow manage to calibrate them well, they  still might be assigning higher confidence to the wrongly predicted samples than the correctly predicted ones. If the minority class samples (positives in the above example) are as important as the majority ones, this  behaviour raises concerns relating to their reliability. Analysing and quantifying such behaviour is necessary to complement our understanding in terms of the reliability of neural networks. In the next section, we discuss this aspect as well.

\begin{figure}
    \centering
    \includegraphics[width=0.3\textwidth]{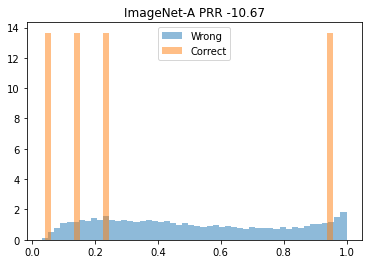}
        \includegraphics[width=0.3\textwidth]{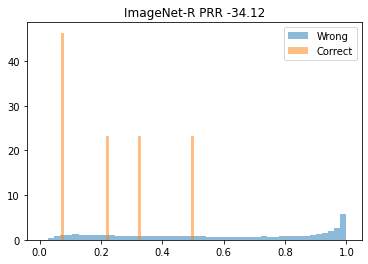}
    \includegraphics[width=0.3\textwidth]{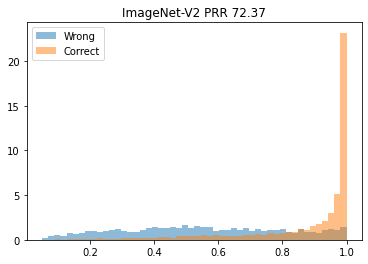}

    \caption{From left to right: the distribution of the confidence values for wrong and right samples for ViT-L/16 on ImageNet-A, ImageNet-R and ImageNet-V2. As it can be seen, in several cases wrong samples are given higher confidence than correct samples (PRR $<$ 0). In cases when PRR $>$ 0, some wrong samples are still given higher confidences (similar to correct samples), but to a lesser extent.}
    \label{fig:hist_prr}
\end{figure}

\subsection{Misclassified Input Detection}
\label{sec:misclassification}

One of the tasks a reliable classifier should be good at is to reject samples on which they are likely to be wrong. This particular task did not receive much attention in the Transformers vs CNNs comparisons performed by the existing literature.  Several ways to evaluate a model at this task are available (e.g. metrics based on ROC \citep{ROCMisclass} or Rejection-Accuracy curves \citep{FumeraRA, NaturalAdversarialExamples}), however, it has already been observed that these metrics favour models that have higher test accuracy \citep{Condessa2015-PerformanceMeasuresRejection, Malinin2019-EnsembleDistillationDirichlet}. A recently proposed metric that allows comparison of different models in this aspect, agnostic to their individual accuracy, is Prediction Rejection Ratio (PRR) \citep{Malinin2019-EnsembleDistillationDirichlet}. The sign of this metric is an indication of whether a model tend to provide lower confidence to correctly classified samples and higher to wrongly classified ones or not. 
The PRR ranges from -1 to 1. It is 0 if the rejection choice is performed at random, negative if the network is more confident on misclassified samples than on correctly classified ones, and positive viceversa. The optimal value of 1 is achieved when the classifier rejects only the misclassified samples while rejecting the most uncertain ones.

For instance, consider Figure \ref{fig:hist_prr} where we show the distribution of the confidence values for samples that a ViT-L/16 wrongly and correctly classified on a few datasets. As observed, in many cases the network is more confident on wrongly classified samples than on the correctly classified ones. This is captured by the sign of PRR (reported in \%). However, the corresponding miscalibration error values as shown in Table \ref{tab:imagenet1k_ds} are particularly low (especially on ImageNet-R). Therefore, as discussed in Section~\ref{sec:calibration}, low miscalibration solely can be misleading in providing a deep understanding of the reliability of different models.

	\begin{table}
		\centering 
\resizebox{0.95\textwidth}{!}{
	\begin{tabular}{c|c|cccc}
		\toprule
		
		 &  \multicolumn{1}{c|}{\textbf{In-Distribution}}  & \multicolumn{4}{c}{\textbf{Domain-Shift}}  \\

		&  \multicolumn{1}{c|}{\textbf{ImageNet-1K (Test)}} & \multicolumn{1}{c}{\textbf{ImageNet-A}} &
		\multicolumn{1}{c}{\textbf{ImageNet-R}} &
		\multicolumn{1}{c}{\textbf{ImageNet-SK}} &
		 \multicolumn{1}{c}{\textbf{ImageNet-V2}}\\
\hline 
		 & \multicolumn{5}{c}{\textbf{PRR} ($\uparrow$)} \\

\hline 

		\hline

BiT-R50x1 &68.38& 54.90 & -25.60 & 58.70 & 63.13\\
BiT-R50x3 &67.61& 28.58 & -42.72 & 60.25 & 64.09\\
BiT-R101x1& 69.82& -0.42 & -25.94 & 60.08 & 64.60\\
BiT-R101x3& 68.93& 29.50 & -34.52 & 60.13 & 65.00\\
BiT-R152x2& 68.03& 31.56 & -35.12 & 59.26 & 63.26\\
BiT-R152x4& 67.00& \textbf{92.04} & -46.05 & 59.34 & 61.48\\
\hline 
ConvNeXt-B& 73.43& 16.03 & -39.91 & 67.44 & 69.84\\
ConvNeXt-L& 73.48& 40.56 & -23.60 & 69.03 & 69.50\\
ConvNeXt-XL& \underline{74.37} & 35.96 & \underline{-19.32} & \underline{69.29} & 70.07\\

\hline 
\hline 
ViT-B16& 74.17& 11.54 & -46.01 & 63.94 & \underline{70.51}\\
ViT-L16& \textbf{76.03}& -10.67 & -34.12 & \textbf{69.79} & \textbf{72.37}\\
\hline 
Swin-B &72.04& 32.65 & -32.95 & 64.23 & 67.35\\
Swin-L &72.89& \underline{56.54} & \textbf{36.53} & 63.52 & 68.49\\

\bottomrule
	\end{tabular}%

    }
    \caption{\textbf{Misclassification detection} results using the PRR (\%) metric.
}
    \label{tab:rrr_ImageNet}
\end{table}

{\bf Comparing Transformers and CNNs}
 As it can be seen form Table \ref{tab:rrr_ImageNet}, in in-distribution ViT-L/16 is the best model, immediately followed by ConvNeXt-XL. ViT-B/16 slightly outperforms ConvNeXt-B and L, which in turn outperform Swin-B and L. On ImageNet-A, the best model is BiT-R152x4, with a significant margin with respect to any other model. The second best model is Swin-L, and the third best is BiT-R50x1. On ImageNet-R, the only model with positive PRR is Swin-L, and the models with highest negative PRR are ConvNeXt-XL and L, followed by BiT-R50x1 and 101x1. On ImageNet-Sketches ViT-L/16 and ConvNeXt-XL perform comparably, immediately followed by ConvNeXt-L and B. On ImageNetV2, ViT-L/16 is the best model, immediately followed by ViT-B/16 and all the ConvNeXts. 

\begin{summary}[title style={colback=grey},colback=white]
\begin{itemize}
    \item[$\circ$] No single model is the winner in detecting misclassified samples.
    \item[$\circ$] The fact that several models are severely overconfident and wrong on ImageNet-R (PRR) while showing low  calibration errors indicate that the calibration analysis should be complemented with experiments such as misclassification detection to understand their reliability.
\end{itemize}

\end{summary}

\section{Concluding Remarks}
We performed an extensive analysis comparing current state-of-the-art Transformers and CNNs. With simple experiments, we have shown that Transformers, just like CNNs, are vulnerable to picking spurious or simple discriminative features in the training set instead of focusing on robust features that generalise under covariate shift conditions. Therefore, the presence of the self-attention mechanism might not be facilitating learning more complex and robust features. To show it is not even necessary, we observed that ConvNeXt models exhibit even superior robustness with respect to current Transformers without leveraging the self-attention mechanism in a few cases. We also conducted an in-depth analysis about the out-of-distribution, calibration, and misclassification detection properties of these models. We hope that our work will encourage development of modules within Transformers and CNNs that can avoid various biases. Additionally, our analysis in Appendix~\ref{sec:capacity} regarding the lack of reliable metrics to quantify a model's capacity to open new avenues for future work.




\section*{Acknowledgements}
This work is supported by the UKRI grant: Turing AI Fellowship EP/W002981/1 and EPSRC/MURI grant: EP/N019474/1. We would  like to thank the Royal Academy of Engineering and FiveAI. Francesco Pinto's PhD is funded by the European Space Agency (ESA). PD would like to thank Anuj Sharma and Kemal Oksuz for their comments on the draft.

\clearpage
%
%

\bibliographystyle{splncs04}
\bibliography{eccv2022submission}

\newpage
\appendix 
\section{Additional experimental details}
\label{sec:additionalResults}
\subsection{About the evaluation metrics}
All metrics are reported in percentage terms. The out-of-distribution detection metrics leverage the 
entropy. For the misclassification detection tasks, we use the confidence score (i.e. the maximum probability of the softmax) as uncertainty metric, as we find it to be the most effective for the task.

\subsection{The impact of the input preprocessing pipeline}

	\begin{table}
		\centering
\resizebox{\textwidth}{!}{

	\begin{tabular}{c|ccc|ccc|ccc|ccc|ccc|c}
		\toprule
		 & \multicolumn{3}{c|}{\textbf{Clean Data}}  & \multicolumn{12}{c|}{\textbf{Domain-Shift}} & \multicolumn{1}{c}{\textbf{OOD}}  \\

		& \multicolumn{3}{c|}{\textbf{ImageNet-1K (Test)}} & \multicolumn{3}{c|}{\textbf{ImageNet-R}} &
		\multicolumn{3}{c|}{\textbf{ImageNet-A}} &
		\multicolumn{3}{c|}{\textbf{ImageNet-V2}} &
		 \multicolumn{3}{c|}{\textbf{ImageNet-Sk}} & \multicolumn{1}{c}{\textbf{ImageNet-O}}  \\

		  & \textbf{Acc} ($\uparrow$) &  \textbf{ECE} ($\downarrow$) & \textbf{AdaECE} ($\downarrow$)  &
		  \textbf{Acc} ($\uparrow$) &  \textbf{ECE} ($\downarrow$) & \textbf{AdaECE} ($\downarrow$) & \textbf{Acc} ($\uparrow$) &  \textbf{ECE} ($\downarrow$) & \textbf{AdaECE} ($\downarrow$) & \textbf{Acc} ($\uparrow$) &  \textbf{ECE} ($\downarrow$) & \textbf{AdaECE} ($\downarrow$) & \textbf{Acc} ($\uparrow$) &  \textbf{ECE} ($\downarrow$) & \textbf{AdaECE} ($\downarrow$)  & \textbf{AUROC} ($\uparrow$)   \\

\hline 
BiT-R50x1              & 80.05 & 1.58 & 1.76 & 38.98 & 10.01 & 10.01 & 26.89 & 19.73 & 19.67 & 67.98 & 1.75 & 1.74 & 24.72 & 18.39 & 18.39 & 67.01   \\
BiT-R50x3              & 83.59 & 2.65 & 2.51 & 47.25 & 8.51 & 8.51 & 46.72 & 11.66 & 11.63 & 72.36 & 6.30 & 6.08 & 32.81 & 19.00 & 19.00 & 77.99   \\
BiT-R101x1             & 82.04 & 1.16 & 1.06 & 43.65 & 7.49 & 7.49 & 38.32 & 15.79 & 15.72 & 70.97 & 4.33 & 4.29 & 29.10 & 18.28 & 18.28 & 73.62   \\
BiT-R101x3             & 84.19 & 3.78 & 3.72 & 50.14 & 9.10 & 9.10 & 53.12 & 10.90 & 10.93 & 73.36 & 7.85 & 7.71 & 36.29 & 21.15 & 21.15 & 80.44   \\
BiT-R152x2             & 84.17 & 2.96 & 2.71 & 51.02 & 8.51 & 8.51 & 52.97 & 10.37 & 10.13 & 73.46 & 6.30 & 6.12 & 36.96 & 19.22 & 19.22 & 80.72   \\
BiT-R152x4             & 84.49 & 6.28 & 6.26 & 54.06 & 11.50 & 11.50 & 58.52 & 12.17 & 12.14 & 74.36 & 10.94 & 10.94 & 41.17 & 25.47 & 25.47 & 85.58   \\
\hline 

ConvNeXt-B         & 85.53 & 2.87 & 2.82 & 62.46 & 2.57 & 2.51 & 52.63 & 8.28 & 8.31 & 75.43 & 2.91 & 2.78 & 48.62 & 8.87 & 8.86 & 85.72  \\
ConvNeXt-L         & 86.29 & 2.27 & 2.34 & 64.57 & 3.00 & 3.08 & 58.23 & 7.57 & 7.26 & 76.77 & 3.72 & 3.85 & 50.06 & 10.31 & 10.31 & 89.07  \\
ConvNeXt-XL        & 86.58 & 2.40 & 2.29 & 66.01 & 2.92 & 2.90 & 61.11 & 7.54 & 7.21 & 77.20 & 4.00 & 4.24 & 52.67 & 11.15 & 11.15 & 90.04  \\
\hline 
\hline 
ViT-B/16           & 77.85 & 1.39 & 1.38 & 43.09 & 5.28 & 5.28 & 23.31 & 23.51 & 23.51 & 65.94 & 4.67 & 4.53 & 18.33 & 12.74 & 12.74 & 79.93  \\
ViT-L/16            & 84.33 & 1.72 & 1.70 & 61.75 & 2.88 & 2.88 & 46.36 & 12.55 & 12.39 & 74.15 & 5.52 & 5.43 & 46.21 & 10.56 & 10.56 & 90.63  \\
\hline 

Swin-B            & 84.81 & 8.52 & 8.52 & 59.81 & 2.11 & 2.14 & 49.88 & 8.57 & 8.40 & 75.07 & 5.11 & 5.06 & 45.43 & 7.50 & 7.50 & 83.94  \\
Swin-L            & 85.95 & 5.65 & 5.65 & 64.44 & 2.29 & 2.19 & 58.96 & 6.82 & 6.83 & 76.49 & 3.24 & 3.02 & 49.06 & 8.73 & 8.72 & 87.66  \\
\hline 
\hline 
\multicolumn{17}{c}{Fine-tuned at resolution 384$\times$384} \\
\hline 

ConvNeXt-B-384     & 86.51 & 3.16 & 3.15 & 64.12 & 3.36 & 3.47 & 63.25 & 7.70 & 7.58 & 77.03 & 2.49 & 2.65 & 50.31 & 7.84 & 7.84 & 87.11  \\
ConvNeXt-L-384     & 87.14 & 2.39 & 2.38 & 66.09 & 3.27 & 3.16 & 66.52 & 7.01 & 6.90 & 77.97 & 3.51 & 3.31 & 51.68 & 9.60 & 9.60 & 90.45  \\
ConvNeXt-XL-384    & 87.45 & 2.37 & 2.49 & 67.24 & 3.22 & 3.35 & 69.59 & 7.28 & 7.29 & 78.34 & 3.03 & 2.87 & 53.80 & 8.69 & 8.67 & 91.12  \\
\hline 

ViT-B16-384        & 79.43 & 1.53 & 1.60 & 40.62 & 6.49 & 6.49 & 33.63 & 17.46 & 17.46 & 68.37 & 4.45 & 4.45 & 14.54 & 15.75 & 15.75 & 81.75  \\
ViT-L16-384        & 85.80 & 2.09 & 1.93 & 63.26 & 3.31 & 3.31 & 63.07 & 6.11 & 5.86 & 76.47 & 5.29 & 5.25 & 46.10 & 12.38 & 12.38 & 92.42   \\
\hline 

SWIN-B-384         & 86.29 & 6.78 & 6.78 & 63.41 & 2.29 & 2.28 & 62.20 & 6.57 & 6.52 & 76.65 & 3.80 & 3.83 & 48.43 & 8.43 & 8.43 & 86.46  \\
SWIN-L-384         & 87.01 & 6.58 & 6.58 & 66.40 & 3.40 & 3.50 & 67.92 & 7.37 & 7.29 & 77.51 & 3.89 & 3.79 & 50.29 & 7.62 & 7.62 & 89.25  \\
\bottomrule
	\end{tabular}%

    }
    \caption{Analogous of Tables \ref{tab:imagenet1k_clean} and Table \ref{tab:imagenet1k_ds} but using the prepocessing pipeline suggested suggested by the timm library for each model. The conclusions of the main paper do not change. 
}
    \label{tab:imagenet1k_timm}
\end{table}

	\begin{table}
		\centering 
\resizebox{\textwidth}{!}{
	\begin{tabular}{c|c|c|c|c|c}
		\toprule
		 & \multicolumn{1}{c|}{\textbf{Clean Data}}  & \multicolumn{4}{c}{\textbf{Domain-Shift}}  \\
		
		
		& \multicolumn{1}{c|}{\textbf{ImageNet-1K (Test)}} & \multicolumn{1}{c}{\textbf{ImageNet-A}} &
		\multicolumn{1}{c}{\textbf{ImageNet-R}} &
		\multicolumn{1}{c}{\textbf{ImageNet-SK}} &
		\multicolumn{1}{c}{\textbf{ImageNet-V2}}  \\
\hline
		 & \multicolumn{5}{c}{\textbf{PRR} ($\uparrow$)} \\
		\hline

BiT-R50x1 &72.48& 23.31 & -25.84 & 56.68 & 68.08 \\
BiT-R50x3 &73.41& -19.39 & -8.67 & 62.41 & 67.39 \\
BiT-R101x1 &74.04& 16.70 & -22.27 & 60.64 & 68.12 \\
BiT-R101x3 &73.39& 15.32 & \textbf{-8.64} & 62.54 & 66.61 \\
BiT-R152x2 &73.24& \underline{48.97} & -20.76 & 61.36 & 66.81 \\
BiT-R152x4 &71.82& 23.89 & -35.15 & 62.15 & 64.54 \\ 
\hline 
ConvNeXt-B &73.43& 16.03 & -39.91 & 67.48 & 69.84 \\ 
ConvNeXt-L &73.48& 40.56 & -23.60 & 69.04 & 69.50 \\ 
ConvNeXt-XL & \textbf{74.36}& 35.96 & -19.32 & \underline{69.29} & 70.07 \\ 

\hline 
\hline 
ViT-B16 &74.12& \textbf{49.46} & -33.53 & 64.59 & 70.89 \\ 
ViT-L16 &76.24& 5.92 & -31.09 & \textbf{69.70} & \textbf{72.61} \\ 

\hline 
Swin-B &71.99& 17.10 & -16.70 & 63.98 & 67.61 \\ 
Swin-L &72.70& -10.78 & -25.43 & 63.83 & 68.91 \\ 

\hline 
\hline 
& \multicolumn{5}{c}{Fine-tuned at resolution 384$\times$384} \\
\hline 
ConvNeXt-B-384 &74.06& 36.79 & -22.39 & 67.37 & 68.75 \\ 
ConvNeXt-L-384 &74.12& 32.74 & \textbf{-10.88} & 68.47 & 69.16 \\ 
ConvNeXt-XL-384 &74.71& 55.16 & -12.21 & \underline{69.05} & 70.27 \\ 
\hline 

ViT-B/16-384 &74.35& 46.47 & -32.97 & 66.18 & 71.13 \\ 
ViT-L/16-384 &\textbf{76.89}& -9.48 & -20.06 & \textbf{69.41} & \textbf{72.76} \\ 
\hline 

Swin-B-384 &72.53& \textbf{69.93} & -42.21 & 63.73 & 67.58 \\ 
Swin-L-384 &71.73& 27.26 & -17.02 & 63.04 & 66.71\\

\bottomrule
	\end{tabular}%

    }
    \caption{Analogous of Table \ref{tab:rrr_ImageNet} but using the preprocessing pipeline suggested by the timm library for each model. The conclusions of the main paper do not change. 
}
    \label{tab:imagenet1k_aurra_timm}
\end{table}

\textbf{The standard pre-processing pipeline} For the results reported in the main paper, we apply the standard ImageNet-1K  test pre-processing pipeline: we first rescale the image at resolution $256 \times 256$ then extract the center crop of $224 \times 224$ and normalise with respect to the mean and standard deviation of the training set.

\textbf{Model-specific pre-processing pipelines} However, it should be noticed that the timm library suggests using a different pre-processing pipeline for each architecture.  We do not follow this procedure for the results in the main paper as fine-tuning the test pre-processing pipeline hyperparameters would require a cross-validation procedure to not overfit the test set and we want to have a fair comparison using the same evaluation procedure for all models. We report the results applying the timm-proposed preprocessing pipelines in Tables \ref{tab:imagenet1k_timm} and \ref{tab:imagenet1k_aurra_timm}. All the conclusions drawn in the main paper about ConvNeXts, ViTs and SwinTranformers do not change. The only case in which altering the pipeline dramatically changes the performance is on BiT models. With respect to the performance with the default pre-processing pipeline, BiT models become: 
\begin{itemize}
    \item[$\circ$] \textbf{significantly more accurate on in-distribution data}. For instance,  BiT-R152$\times$4's accuracy jumps from 78.16\% to 84.49\%.
    \item[$\circ$] \textbf{significantly more accurate on covariate shifted inputs}. Particularly remarkable is the improvement when exposed to ImageNet-A. For instance, the accuracy of BiT-R50$\times$1 jumps from 10.97 to 38.98 (which renders the smallest BiT model better performing than ViT-B/16!). Similarly, larger capacity BiT models can outperform ViT-L/16 and BiT-R152$\times$4 is as competitive as the top-performing transformer (Swin-L). It is important to recall that ImageNet-A samples were selected to produce low accuracy on ResNets. This selection bias obviously makes comparisons between ResNets and any other architecture unfair. However, already changing the pre-processing pipeline at test time is enough to significantly weaken the adversarial effectiveness of the selection process on ResNet inspired architectures. Similarly, on other data-shift datasets, BiTs become extremely more competitive, and can outperform or be almost comparable to smallest transformer variants in many cases. 
    \item[$\circ$]   \textbf{significantly better at out-of-distribution detection} (e.g. the minimum gap between BiT models and ViT-L/16 passes from almost 11\% to less than 7\%)  
    \item[$\circ$] \textbf{significantly more calibrated on both in-domain and covariate shifted inputs}. (e.g. the ECE is approximately halved in most cases on in-distribution data)
    \item[$\circ$] \textbf{significantly better at performing in-domain misclassification detection and most distribution-shift experiments}. It increases (in most cases) on ImageNet-R, ImageNet-SK and ImageNet-V2. On ImageNet-A the performance decreases. This is another interesting case in which the calibration and misclassification detection provide complementary information: while the calibration error decreases on ImageNet-A, the misclassification detection performance  gets worse, indicating the problem of being overconfidently wrong becomes more pronounced. 
\end{itemize}

\textbf{Models fine-tuned at resolution 384$\times$384} It should also be noticed that variants fine-tuned at resolution 348$\times$384 exist (see the lower parts of Table \ref{tab:imagenet1k_timm} and Table \ref{tab:imagenet1k_aurra_timm}). These variants generally outperform the variants fine-tuned at lower resolution in terms of accuracy, but generally exhibit worse uncertainty properties. The final conclusions of our paper do not change when considering these variants. Since we could not find BiT checkpoints fine-tuned at this resolution in the timm library, we reported the performance for models fine-tuned at 224$\times$224 to have a fair comparison.

\subsection{The impact of pre-training}
It would be interesting to study the robustness and reliability of models without pre-training on ImageNet-21K. Unfortunately, checkpoints training solely on ImageNet-1K are often not included in the timm library or in general not publicly available, mostly because some of the models considered do not produce good performance if trained from scratch on ImageNet-1K.

For completeness, we report the performance results on ConvNeXt-B/L and Swin-B in Tables \ref{tab:imagenet1k_nopretrain} and \ref{tab:imagenet1k_aurra_nopretrain} . Notice, in this case the out-of-distribution detection results are reported using the negative confidence score as a form of uncertainty, as we find it to be the most effective in this case.

As it can be seen in Table \ref{tab:imagenet1k_nopretrain}, ConvNeXt-B is typically more accurate and better calibrated than Swin-B except on ImageNet-A and ImageNet-V2 (where Swin-B is more calibrated). Swin-B produces better out-of-distribution detection performance. 
As seen in Table \ref{tab:imagenet1k_aurra_nopretrain}, ConvNeXt-L outperforms all other models at misclassification detection except in one case. However, we cannot draw conclusions from these two tables given the lack of comparison with other strong Transformer architectures and CNNs. 

We can however evaluate the difference between with and without pretraining as follows: 
\begin{itemize}
    \item[$\circ$] in all cases, in-distribution accuracy is significantly improved by pre-training
    \item[$\circ$] for ConvNeXt models, the lack of pre-training harms calibration on in-domain data and under covariate shift. For Swin-B, the lack of pretraining improves the calibration on in-distribution data, but harms it under covariate shift. 
    \item[$\circ$] the lack of pretraining significantly damages the out-of-distribution detection performance of all models
    \item[$\circ$] the lack of pretraining harms the misclassification detection performance except in the case of ImageNet-R for ConvNeXt-B and Swin-B. Swin-B performance drops more significantly than ConvNeXt-L models without pretraining.
\end{itemize}

	\begin{table}
		\centering
\resizebox{\textwidth}{!}{

	\begin{tabular}{c|ccc|ccc|ccc|ccc|ccc|c}
		\toprule
		 &  \multicolumn{3}{c|}{\textbf{Clean Data}}  & \multicolumn{12}{c|}{\textbf{Domain-Shift}} & \multicolumn{1}{c}{\textbf{OOD}}  \\

		&  \multicolumn{3}{c|}{\textbf{ImageNet-1K (Test)}} & \multicolumn{3}{c|}{\textbf{ImageNet-R}} &
		\multicolumn{3}{c|}{\textbf{ImageNet-A}} &
		\multicolumn{3}{c|}{\textbf{ImageNet-V2}} &
		 \multicolumn{3}{c|}{\textbf{ImageNet-Sk}} & \multicolumn{1}{c}{\textbf{ImageNet-O}}  \\

		 & \textbf{Acc} ($\uparrow$) &  \textbf{ECE} ($\downarrow$) & \textbf{AdaECE} ($\downarrow$)  &
		  \textbf{Acc} ($\uparrow$) &  \textbf{ECE} ($\downarrow$) & \textbf{AdaECE} ($\downarrow$) & \textbf{Acc} ($\uparrow$) &  \textbf{ECE} ($\downarrow$) & \textbf{AdaECE} ($\downarrow$) & \textbf{Acc} ($\uparrow$) &  \textbf{ECE} ($\downarrow$) & \textbf{AdaECE} ($\downarrow$) & \textbf{Acc} ($\uparrow$) &  \textbf{ECE} ($\downarrow$) & \textbf{AdaECE} ($\downarrow$)  & \textbf{AUROC} ($\uparrow$)   \\

ConvNeXt-B  & 83.73 & 3.33 & 3.43 & 51.72 & 8.18 & 8.14 & 35.79 & 22.55 & 22.51 & 73.69 & 5.55 & 6.30 & 38.27 & 22.78 & 22.78 & 62.64 \\ 
ConvNeXt-L  & 84.16 & 3.86 & 3.95 & 53.93 & 8.50 & 8.47 & 40.54 & 21.42 & 21.40 & 74.01 & 5.74 & 6.47 & 40.14 & 23.40 & 23.40 & 62.68 \\ 
\hline 
Swin-B  & 83.08 & 5.08 & 5.01 & 47.20 & 8.72 & 8.71 & 34.39 & 20.43 & 20.45 & 72.10 & 5.29 & 4.99 & 32.62 & 22.83 & 22.83 & 64.01  \\ 

\bottomrule 
	\end{tabular}%

    }
    \caption{Analogous of Table \ref{tab:imagenet1k_timm}, but checkpoints are not pre-trainined on ImageNet-21K.
}
    \label{tab:imagenet1k_nopretrain}
\end{table}

	\begin{table}
		\centering 
\resizebox{\textwidth}{!}{
	\begin{tabular}{c|c|c|c|c|c}
		\toprule
		 & \multicolumn{1}{c|}{\textbf{Clean Data}}  & \multicolumn{4}{c}{\textbf{Domain-Shift}}  \\
		
		
		& \multicolumn{1}{c|}{\textbf{ImageNet-1K (Test)}} & \multicolumn{1}{c}{\textbf{ImageNet-A}} &
		\multicolumn{1}{c}{\textbf{ImageNet-R}} &
		\multicolumn{1}{c}{\textbf{ImageNet-SK}} &
		\multicolumn{1}{c}{\textbf{ImageNet-V2}}  \\
\hline
		 & \multicolumn{5}{c}{\textbf{PRR} ($\uparrow$)} \\
		\hline

ConvNeXt-B & 70.45& -7.13 & -11.43 & 65.16 & 65.31 \\
ConvNeXt-L & 69.71& 36.21 & -30.27 & 64.04 & 65.93 \\
\hline 
Swin-B & 68.20& 34.16 & -2.43 & 59.58 & 63.18 \\

\bottomrule
	\end{tabular}%

    }
    \caption{Analogous of Table \ref{tab:imagenet1k_aurra_timm}, but checkpoints are not pre-trained on ImageNet-21K.
}
    \label{tab:imagenet1k_aurra_nopretrain}
\end{table}

\section{Further discussion on why the practice of comparing models based on parameter count might be misleading}
\label{sec:capacity}
In this section we provide additional examples explaining why the parameter count is not really a representative of a model's capacity and its generalizability i.e., the ability of a model to capture better approximations of the function underlying the relationship between inputs and outputs that generalise better. 

One might wonder whether ways to quantify this aspect of a model exist. For this reason, we resort to the known complexity measures in the literature and show that these are no better than parameter count for the purpose of comparing models belonging to different families of architectures. This advocates for the need of measures that allow to compare models independently of their kinship.

\subsection{Practical examples indicating why parameter count is not a good proxy to compare model capacity and generalization}

It is important to observe that all the considered models have significantly more parameters than the number of training samples (even when considering ImageNet-21K as training set). Therefore, from a theoretical point of view, all the considered models can interpolate the training set. These models differ in the way in which their learning procedures can leverage the data and the available parameters to learn solutions that generalise better. How overparametrization is related to the extraordinary generalization properties of Neural Networks is still an open area of research, and out of the scope of this paper. Consider the following examples (refer Table~\ref{tab:complexity} for parameter counts):
\begin{itemize}
    \item[$\circ$] consider BiT-R152$\times$2 and BiT-R152$\times$4. It is evident that although the latter has about 4 times the number of parameters of the former, the performance improvements observed in our tables are often marginal. This implies the training procedure is not capable of leveraging the additional number of parameters to boost the performance. It will be interesting if the future literature investigates how much BiT-R152$\times$4 can be pruned before it looses its advantage over BiT-R152$\times$2.
    \item[$\circ$] consider ConvNeXt-B and BiT-R152$\times$4. Although the first contains almost 10$\times$ less parameters than the latter, and both rely on convolutional inductive biases, ConvNeXt-B significantly outperforms BiT-R152$\times$4 almost always. This comparison shows that parameter count is not representative of the generalisation properties of a model even when comparing models sharing convolutional inductive biases. Several other design choices that are often neglected in existing literature comparing the robustness and reliability of Transformers and CNNs (e.g. quantity and types of activations or normalization layers, kernel sizes, proportions between the block sizes etc.)  can greatly influence the ability of a model to produce robust and reliable predictions.    
\end{itemize}

\subsection{Can complexity measures do better than parameter count?}

A natural question that arises from the above observations is whether it is possible to find a measure that quantifies the generalization properties of a model as a function of its input-output behaviour, training dynamics, the properties of the mappings it has learned, or all these combined together. A recent study collected and compared the most popular measures in this regard \citep{FantasticComplexity}. We consider the two most popular ones and show how they cannot be used to compare the generalization properties of models belonging to different families, and therefore, for this purpose, are no more useful than parameter count. 
\begin{itemize}
    \item[$\circ$] Path-Norm \citep{PathNorm}, defined as:
      $$\text{PN} = \sum_i f_{w^2}(\mathbf{1})[i]$$
      
      where $f_{w^2}$ represents a network whose parameters have been squared, $\mathbf{1}$ indicates an input (of adequate shape, in this case we apply the same shape of ImageNet inputs) for which each entry is set to 1,  $f_{w^2}(.)[i]$ represents the logit associated to class i. 
    \item[$\circ$] (logarithm of) Spec-Fro \citep{SpecFro}, defined as:
$$\log \text{SF} = \log \prod_{i=1}^L ||\textbf{W}_i||_2^2 \sum_{i=1}^L \text{srank}(\textbf{W}_i)$$
    where $L$ is the total number of layers in the network, $\textbf{W}_i$ represents the weight matrix of the i-th linear layer, $\text{srank}(\textbf{W}_i)$ represents the stable rank of $\textbf{W}_i$, i.e. $\text{srank}(\textbf{W}_i) = ||\textbf{W}_i||_F^2 / ||\textbf{W}_i||_2^2$~\cite{Sanyal2020-SRN}. The logarithm is taken for numerical stability reasons. 
\end{itemize}

As shown in Table \ref{tab:complexity}, both these metrics are inadequate in comparing models belonging to different families (e.g. Path-Norm and Spec-Fro of BiT are evidently at another scale with respect to those of other models; also, no inter-family consistent sorting based on generalization on the in-domain test set seems to emerge). Also for the same architecture, the behaviour of these metrics is inconsistent when comparing models pre-trained on ImageNet-21K and then fine-tuned on ImageNet-1K with respect to those trained only on ImageNet-1K. For instance, in the case of ConvNeXt these metrics remain almost unchanged, while for Swin-B the change is dramatic. They also produce inconsistent behaviours within a family, for instance, they do not sort based on generalisation properties the ViT-B and L models with patch sizes 16 and 32. For these reasons, for the purposes of this analysis, these metrics are no more useful than the parameter count. Future research should address this issue. 

\begin{table}
		\centering 
\resizebox{0.5\textwidth}{!}{
	\begin{tabular}{cc|cc}

& $\#$ params & Path-Norm  & log-Spec-Fro  \\
\hline

BiT-R50$\times$1    &25& 71.90 & 101.179    \\
BiT-R50$\times$3    &217 & 211.21& 103.10  \\
BiT-R101$\times$1   &44 & 75.43& 197.44 \\
BiT-R101$\times$3   &387 & 224.96& 199.62  \\
BiT-R152$\times$2  &232 & 151.50 & 295.13 \\
BiT-R152$\times$4   &936 & 298.22&296.47\\ 
\hline

ConvNeXt-B    &88 & 0.51 & 2.23   \\
ConvNeXt-L    &196 & 0.75 &55.60  \\
ConvNeXt-XL   &348 & -0.28 & 80.75    \\
\hline
\hline

ViT-B/16     &86 & 0.34 & 46.77 \\
ViT-L/16     &304 & -0.44 &  118.72 \\
ViT-B/32     &88 & 0.17 & 46.20 \\
ViT-L/32     &306 & 0.86 &  118.17 \\

\hline

Swin-B     &87 & 0.04 & 34.11\\
Swin-L     &195 & -0.95 & 84.72\\
\hline
\hline
\multicolumn{4}{c}{Trained on ImageNet-1K only} \\
\hline
ConvNeXt-B    &88 & 0.50 & 2.22   \\
ConvNeXt-L    &196 & 0.76 & 55.60  \\
\hline 
Swin-B     &87 & -314.96 & 381.30\\

\bottomrule
	\end{tabular}%

    }
    \caption{\textbf{Path-Norm and Spec-Fro Complexity measures} for each of the considered models (checkpoints pre-trained on ImageNet-21K and fine-tuned on ImageNet-1K, except for the bottom part of the table
}
    \label{tab:complexity}
\end{table}

\section{Samples of the ImageNet-9 and Cue-Conflict dataset }
\label{sec:sampleimages}
\begin{figure}
    \centering
    \includegraphics[width=0.2\textwidth]{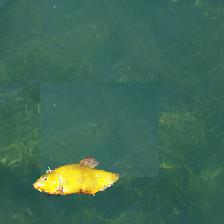}
    \includegraphics[width=0.2\textwidth]{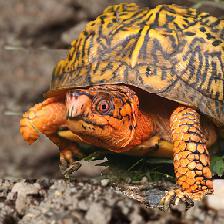}
    \includegraphics[width=0.2\textwidth]{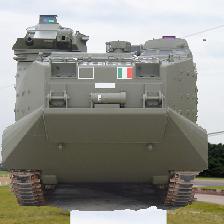}
    \includegraphics[width=0.2\textwidth]{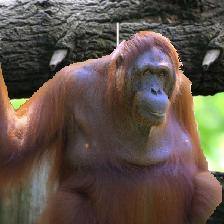}

    \caption{Samples from the ImageNet-9 mixed-same split, in which the foreground of a class is mixed with a background from the same class.}
    \label{fig:mix_same}
\end{figure}

\begin{figure}
    \centering
    \includegraphics[width=0.2\textwidth]{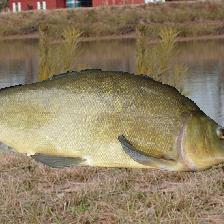}
    \includegraphics[width=0.2\textwidth]{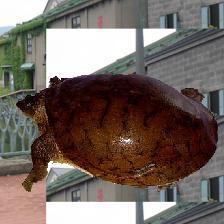}
    \includegraphics[width=0.2\textwidth]{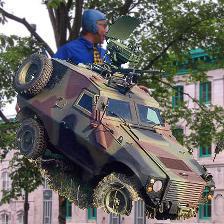}
    \includegraphics[width=0.2\textwidth]{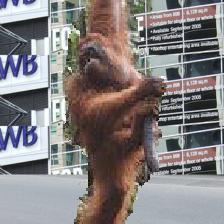}

    \caption{Samples from the ImageNet-9 mixed-random split, in which the foreground of a class is mixed with a background from another class.}
    \label{fig:mixed_random}
\end{figure}

\begin{figure}
    \centering
    \includegraphics[width=0.2\textwidth]{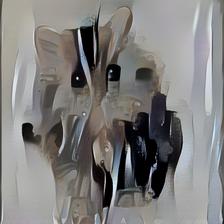}
        \includegraphics[width=0.2\textwidth]{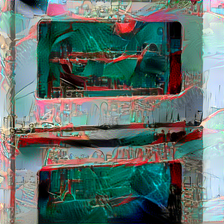}
    \includegraphics[width=0.2\textwidth]{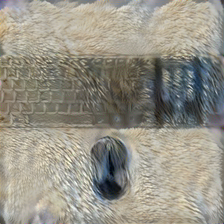}
    \includegraphics[width=0.2\textwidth]{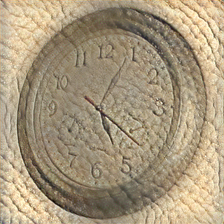}

    \caption{Samples from the Cue-Conflict dataset where style transfer is used to alter the texture of an image using the image from another class as the style source.}
    \label{fig:ccs}
\end{figure}

To provide better context, in Figures \ref{fig:mix_same} and \ref{fig:mixed_random} we show a few  samples from the ImageNet-9 mixed-same and mixed-random splits. In Figure \ref{fig:ccs} we show samples from the Cue-Conflict dataset.

\section{Proof that AUROC is invariant to the choice of positive and negative classes}
\label{sec:auroc_invariant}
Here we provide a simple proof to show that for the binary threshold classifier, AUROC does not vary depending on the choice of positive and negative classes. 
We would like to mention that we do not claim any technical novelty here. This proof is entirely for the purpose of completeness and to theoretically support our empirical findings in Table \ref{tab:ood}. 

Given a classifier $f:\bfx \mapsto \real^k$ and a scoring function $g: \real^k \mapsto \real$ (e.g. entropy), let the binary threshold classifier be such that $g(f(\bfx)) \geq t$ for a given threshold $t$ implies that the sample $\bfx$ belongs to `positive' class, otherwise negative. Therefore, given a dataset with M samples, one could simply sort these samples using the $g(.)$ scores and find an index beyond which all the samples belong to the positive class. Now, let us define $TP$ as the number of true positives (similarly, $FN$, $FP$, and $TP$ are defined). Total number of positive samples can then be calculated as $P = TP + FN$. Similarly, total number of negatives $N = TN + FP$. Using these notations, following rates can be defined
\begin{compactitem}
\item[$\circ$] $\TPR = TP/(TP+FN)$ (True Positive Rate, also called Recall\footnote{Precision = $TP/(TP+FP)$})
\item[$\circ$] $\FPR = FP/(FP+TN)$ (False Positive Rate)
\item[$\circ$] $\TNR = TN/(TN+FP)$ (True Negative Rate) 
\item[$\circ$] $\FNR = FN/(FN+TP)$ (False Negative Rate)
\end{compactitem}

By definition, AUROC is the area under the TPR and FPR curve, where each (TPR($t_i$), FPR($t_i$)) point on the curve is specific to a particular threshold $t_i$ that is used for the classification using the score $g(.)$. 

Let $t_i > t_j$ if $i>j$ then, it is simple to observe that the ROC is a monotonically increasing (not strictly) stair function whose step occur in correspondence of an input $\bfx$ in the dataset. Since the dataset size is fixed (say M samples), one can identify at most M values of $t$ at which the ROC value could increase. Let (TPR($t_i$), FPR($t_i$)) be the element of this ordered set. Since the dataset size is fixed, there are at most $M$ increasing values of the threshold $t$. Then, the area under ROC curve can be obtained as $$\AUROC=\sum_{i=1}^M (b_i - b_{i-1})h_i$$
where, $b_i = FPR(t_i)$ and $h_i = TPR(t_i)$.

Now, let us flip the labels and  $-g(f(\bfx)) \geq t$  is considered as the `positive' class\footnote{Any strictly monotonically decreasing function applied to $g$ will not change what follows.} (note, this way scoring  function reverts the order with which the dataset can be sorted). Let  $TP'$ now denote the true-positive in this new scenario. Similarly, let's apply the same convention to the other elements of the confusion matrix in this scenario. The relation between the confusion matrices of this scenario and the previous is:
\begin{compactitem}
\item[$\circ$] $TP' = TN$
\item[$\circ$] $FP' = FN$
\item[$\circ$] $FN' = FP$
\item[$\circ$] $TN' = TP$
\end{compactitem}
Therefore, 
\begin{align}
\label{eq:tprtofpr}
    \TPR=TP/(TP+FN) = TN'/(TN'+FP') = \TNR' = 1-\FPR', \nonumber \\
    \FPR = FP/(FP+TN) = FN'/(FN'+TP') = \FNR' = 1-\TPR'.
\end{align}
 
The AUROC can now be computed as:
$$\AUROC'=\sum_{i=1}^M (b'_i - b'_{i-1})h'_i.$$ 
It is easy to observe that since the sorting of the samples based on their scoring function values is reversed, harmonising the indexing across these two sorted sets, then $b'_i = 1-h_{M-i}$ and $h'_i = 1 - b_{M-i}$ (applying Eq. (1)). 
 Replacing these values in $\AUROC'$ we obtain
$$ \begin{aligned}
\AUROC' & =\sum_{i=1}^M (1 - h_{M-i} - (1 - h_{M-i+1}))(1-b_{M-i}) \\ &= \sum_{i=1}^M (h_{M-i+1} - h_{M-i})(1-b_{M-i}) = \AUROC
\end{aligned}$$

The last equality is obvious with geometric reasoning: while the $\AUROC$ partitions the ROC with vertical rectangles (one for each step) and summates the area of the rectangles obtained this way, $\AUROC'$ partitions the same ROC stair using horizontal lines (one for each step) and summates the area of these rectangles (which is obviously the same). This is further validated by reasoning geometrically on the mapping implied by the swap of the positive class: this mapping moves the origin of the original space to $(1,1)$, rotates the coordinate axes of 90° anti-clockwise and flips what used to be the horizontal axis orientation. 

Additionally, it is straightforward to observe that when the minority class is sampled multiple times for rebalancing, TPR and FPR are unchanged, therefore, AUROC is unchanged.

\end{document}